\documentclass{article}
\usepackage{spconf}

\usepackage{graphics}
\usepackage{graphicx}
\usepackage{epsfig}

\usepackage{amssymb}
\usepackage{amsthm}

\usepackage{multirow}

\usepackage{subfigure}
\usepackage{graphicx}
\usepackage{amsmath}
\usepackage{amssymb}
\usepackage{color}
\usepackage{arydshln}
\usepackage{rotating}

\title{Fully Automatic Segmentation of Lumbar Vertebrae from CT Images using Cascaded 3D Fully Convolutional Networks }
\name{Rens Janssens, Guodong Zeng and Guoyan Zheng}

\address{Institute for Surgical Technology \& Biomechanics, University of Bern, Bern, Switzerland. \\
				Contact: \texttt{guoyan.zheng@istb.unibe.ch}
				}

\begin{document}
%\ninept
%
\maketitle
\begin{abstract}
We present a method to address the challenging problem of segmentation of lumbar vertebrae from CT images acquired with varying fields of view. Our method is based on cascaded 3D Fully Convolutional Networks (FCNs) consisting of a localization FCN and a segmentation FCN. More specifically, in the first step we train a regression 3D FCN (we call it ``LocalizationNet") to find the bounding box of the lumbar region. After that, a 3D U-net like FCN (we call it ``SegmentationNet") is then developed, which after training, can perform a pixel-wise multi-class segmentation to map a cropped lumber region volumetric data to its volume-wise labels. Evaluated on publicly available datasets, our method achieved an average Dice coefficient of 95.77 $\pm$ 0.81\% and an average symmetric surface distance of 0.37 $\pm$ 0.06 mm.
\end{abstract}


%===========================================================================
\section{Introduction}
\label{sec:intro}
%===========================================================================
In clinical routine, lower back pain (LBP) caused by spinal disorders is reported as common reason for clinical visits \cite{sports_medicine}. Both computed tomography (CT) and magnetic resonance (MR) imaging technologies are used in computer assisted spinal diagnosis and therapy support systems. MR imaging becomes the preferred modality for diagnosing various spinal disorders such as degenerative disc diseases, spindylolisthesis and spinal stenosis due to its excellent soft tissue contrast and no ionizing radiation \cite{Emch_SR_2011}. However, when it comes to spine trauma patients with moderate or high risk, CT is superior to all other imaging modalities in the detection of vertebral fractures and unstable injuries \cite{Parizel_EUJ_2010}. 

An accurate segmentation of individual vertebrae from CT images are important for many clinical applications. After segmentation, it is possible to determine the shape and condition of individual vertebrae. Segmentation can also assist early diagnosis, surgical planning and locating spinal pathologies like degenerative disorders, deformations, trauma, tumors and fractures. Most computer-assisted diagnosis and planning systems are based on manual segmentation performed by physicians. The disadvantage of manual segmentation is that it is time consuming and the results are not really reproducible because the image interpretations by humans may vary significantly across interpreters. 

In this paper, we address the challenging problem of automatic segmentation of lumbar vertebrae from 3D CT images acquired with varying fields of view (FOV), which is usually solved with a two-stage method consisting of a localization stage followed by a segmentation stage \cite{Chu_PLOSONE_2015}. The localization aims to identify each lumbar vertebra, where segmentation handles the problem of producing binary labeling of a given 3D image. For vertebra localization, there exist both semi-automatic methods and fully automatic methods~\cite{Yao_CMIG_2016}. For vertebra segmentation, both 2D image-based methods and 3D image-based methods~\cite{Klinder2009,Stern2011,Ibragimov2014,comp_korez2} are introduced before. These methods can be roughly classified as statistical shape model or atlas based methods, and graph theory (GT) based methods. The multiple center milestone study of clinical vertebra segmentation as presented in \cite{Yao_CMIG_2016} summarized the performance of several state-of-the-art vertebra segmentation algorithms on CT scans. 

\begin{figure*}[t]
\centering
\includegraphics[width=16.5cm]{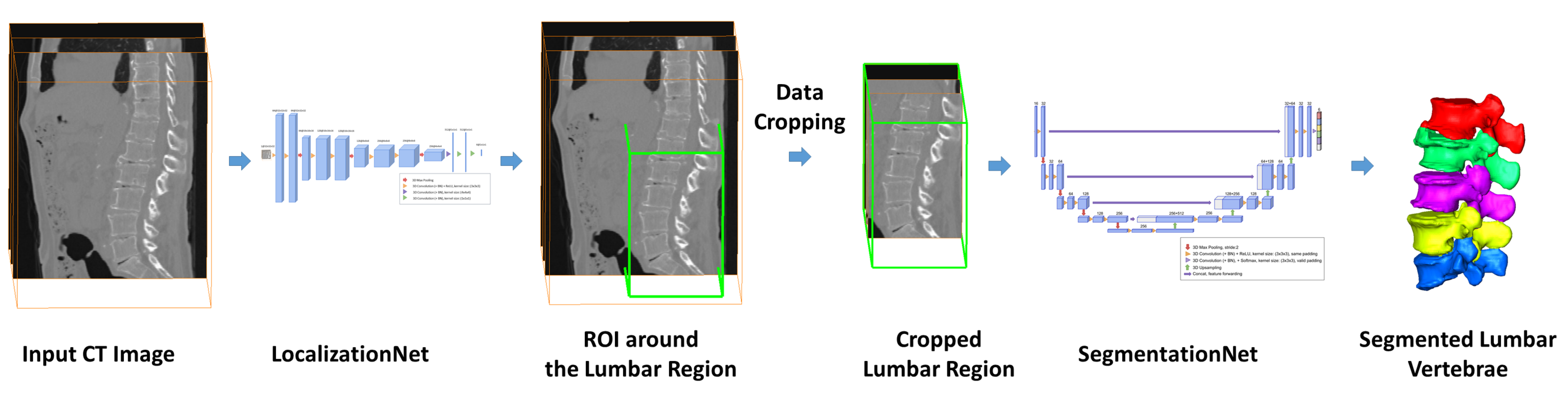}
\caption{ A schematic view of the present approach. }
\label{fig:pipeline}
\end{figure*}

Recently, machine learning-based methods have gained more and more interest in the medical image analysis community. Most of these methods are based on ensemble learning principles that can aggregate predictions of multiple classifiers and demonstrate superior performance in various challenging medical image analysis problems \cite{Chu_PLOSONE_2015}. For example, Michael Kelm et al. \cite{MichaelKelm_MedIA_2013} proposed to detect spine from CT or MR images using iterative marginal space learning. Zhan et al. \cite{Zhan_MICCAI_2012} presented a hierarchical strategy and local articulated model to detect vertebrae and discs from 3D MR images. Due to the successful applications of Random Forest (RF) regression and classification for localization and segmentation of orans from 3D CT/MR data, such a technique has been used for automatic localization and segmentation of vertebrae from CT/MR images \cite{Chu_PLOSONE_2015,Glocker_MICCAI_2012}. 

More recently, with the advance of deep learning techniques \cite{long2015fully,ronneberger2015u,3DUNET_2016}, many researchers have proposed deep learning based methods for automatic localization and segmentation of vertebrae from CT images. For example, Chen et al. \cite{Chen_MICCAI_2015} proposed a method for automatically locating and identifying vertebrae in 3D CT volumes by exploiting high level feature representation with deep convolutional neural networks (CNN). To solve the same task, a different approach was presented by Suzani et al. \cite{Suzani_MICCAI_2015} where they parametrized the vertebral localization problem as a multi-variate non-linear regression. They then used deep feed-forward neural network with hand-crafted features to regress displacements between the center of each vertebral body and reference voxels which were selected using Canny edge detector. The final estimation of the center for each vertebral body was then obtained by using an adaptive kernel density estimation method. This idea was later extended by Sekuboyina et al. \cite{Sekuboyina_arXiv_2017} to develop a localization-segmentation approach for automatic segmentation of lumbar vertebrae.  More specifically, instead of localizing each individual vertebra, they proposed to use a multi-layered perceptron (MLP) with hand-crafted features to perform non-linear regression for locating the lumber region. After that, a 2D U-net like Fully Convolutional Networks (FCN) was used to segment all sagittal slices in order to get a volumetric segmentation.

Inspired by \cite{Sekuboyina_arXiv_2017}, in this paper we proposed a method to automatically segment lumbar vertebrae from 3D CT images using cascaded 3D FCNs consisting of a localization FCN and a segmentation FCN (see Fig. \ref{fig:pipeline} for an overview). More specifically, in the first step we train a regression 3D FCN (the LocalizationNet in Fig. \ref{fig:pipeline}) to find a bounding box which defines the region of interest (ROI) of the lumbar region. After that, a 3D U-net like FCN (the SegmentationNet in Fig. \ref{fig:pipeline}) is then developed, which after training, can perform a pixel-wise multi-class segmentation to map a cropped lumber region volumetric data to its volume-wise labels.

The paper is organized as follows. In the next section, we will describe the method. Section 3 will present the experimental results, followed by discussions and conclusions in Section 4.

\section{Method}
\label{sec:Method}
Although we require all CT data used in our study containing lumbar region, the FOV of each scan varies from data to data. Some may include vertebrae as high as T1 vertebra. Thus, it is important to first localize the lumbar region and then to develop a segmentation method to segment each individual lumbar vertebra. This has motivated us to develop cascaded FCNs consisting of a LocalizationNet and a SegmentationNet as shown in Fig. \ref{fig:pipeline}. Below we first present the LocalizationNet, followed by a description of the SegmentationNet. 

\begin{figure}[t]
\centering
\includegraphics[width=8.5cm]{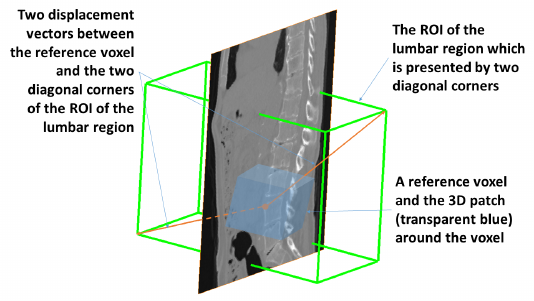}
\caption{ The lumbar region localization problem is formulated as a regression problem. The target of the regression is the two vectors from a reference voxel to the diagonal corners of the ROI of the lumbar region. }
\label{fig:localization}
\end{figure}

\subsection{Lumbar Region Localization}
Inspired by previous work \cite{Suzani_MICCAI_2015,Sekuboyina_arXiv_2017}, we also formulate the lumbar region localization as a multi-variate regression problem. We use a rectangular box to represent the ROI of the lumbar region for each data, which can be represented by two diagonal corners of the rectangular box. The target of the regression is then the two relative displacement vectors between a reference voxel and the two diagonal corners as shown in Fig. \ref{fig:localization}. Following \cite{Suzani_MICCAI_2015}, we used Canny edge detector to select voxels with high edge responses as the reference voxels in both training and testing stages. Unlike previous work, where hand-crafted features computed from a 3D patch sampled around each reference voxel are used to regress the target, here we propose to directly regress the target from a sampled 3D patch using a deep FCN, where the features are automatically learned from the data.     

\begin{figure}[t]
\centering
\includegraphics[width=8.5cm]{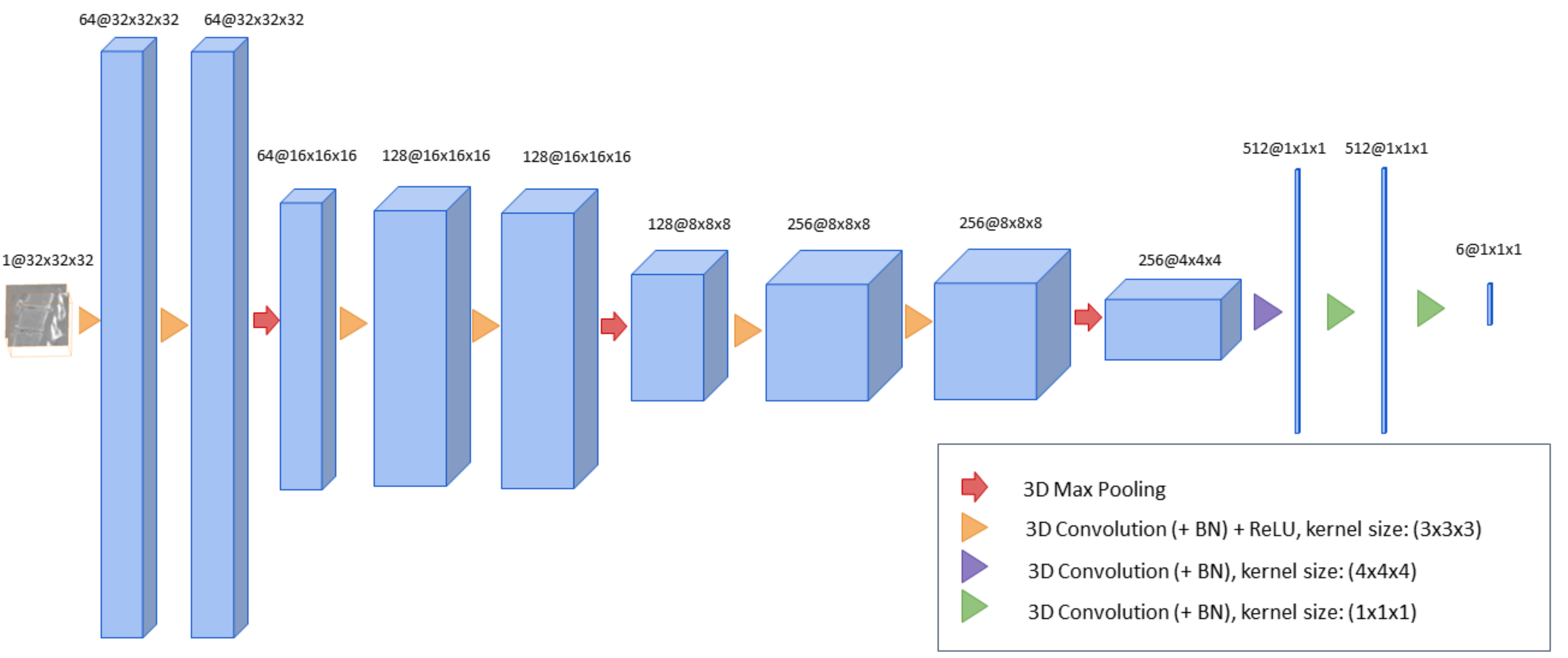}
\caption{ Architecture of the LocalizationNet, which is a 3D regression FCN. The digits above each block take the format as ``number of feature maps@data size". }
\label{fig:localizationnet}
\end{figure}

To this end, we developed a localization net to automatically regress the two target displacement vectors from a sampled 3D patch. The architecture of the LocalizationNet is shown in Fig. \ref{fig:localizationnet}. It consists of three repetitions of two 3D convolutional layers followed by one maximum pooling layer. The convolutional layers have a kernel size of $3 \times 3 \times 3$ and use ``same" padding so that the output size of the convolutional layer is the same as the input size. After each convolutional layer, there is a batch normalization layer (BN) \cite{ioffe2015batch} and Rectified linear unit (ReLU) activation function. The max pooling has a stride of two and halves the spatial dimensionality. The input to the LocalizationNet is a 3D patch of size $32 \times 32 \times 32$. After the three repetitions, the size of the feature maps is $4 \times 4 \times 4$ and is resized with another convolutional layer with a kernel size of $4 \times 4 \times 4$ to the size of $1 \times 1 \times 1$ with 512 features. This process is followed by two $1 \times 1 \times 1$ convolutions. The last convolutional layer reduces the number of output channels to the desired 6 values (representing the two displacement vectors to the two corners).

\textbf{Training}. The LocalizationNet was trained in two rounds. In the first round, the mean squared error loss (L2) is optimized with the Adam solver. In the second round, the Intersection of Union (IoU) loss as introduced in \cite{unitbox} was used to train the pretrained (from round one) network, which has been shown to have better performance than L2 loss for object detection. The IoU could not be used at the beginning yet because for that, the corners have to be estimated in a reasonable accuracy already, otherwise there is no intersection between the estimated ROI and the ground truth ROI.

\textbf{Testing}. During testing, the LocalizationNet can predict two displacement vectors from the center of each sampled 3D patch to respectively the two corners of the ROI . After that, we used kernel density estimation method \cite{KDE_AS_2010} to obtain a density function for all the voxel votes for each corner of the ROI. The global maximum of this density function is considered as the predicted location of the associated corner. 

\subsection{Lumbar Vertebrae Segmentation}
\begin{figure}[t]
\centering
\includegraphics[width=8.5cm]{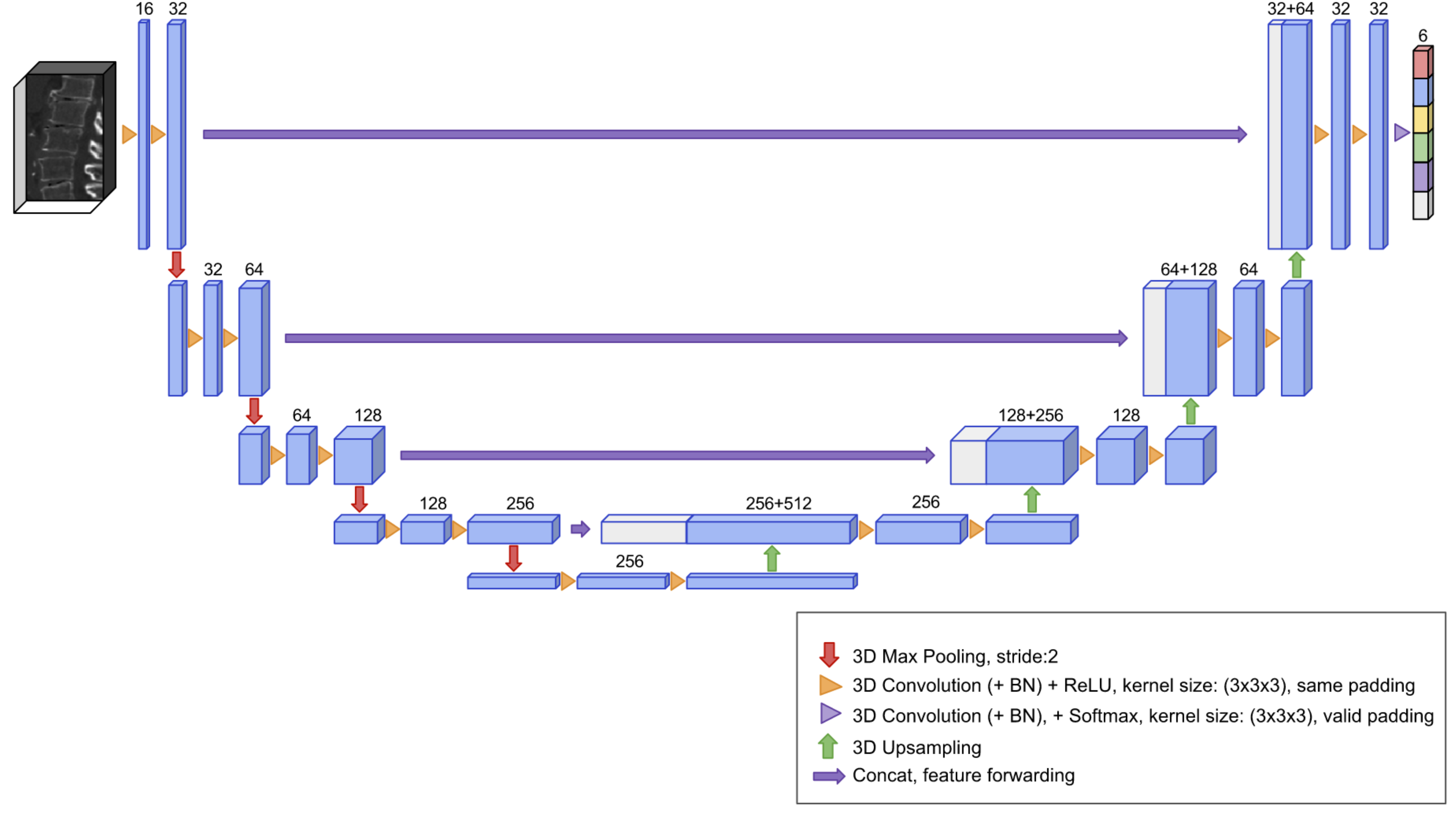}
\caption{ Architecture of the U-net like SegmentationNet. The number above each block indicates the number of feature maps. }
\label{fig:segmentationnet}
\end{figure}

The estimated corners of the ROI will allow us to extract the lumbar region from an input CT data. The goal of this stage is then to segment each individual lumbar vertebra from the cropped data. To this end, we developed a segmentation net to conduct multi-class segmentation of the cropped lumbar region data. A 3D U-net \cite{3DUNET_2016} like FCN was adopted here for our purpose. Fig. \ref{fig:segmentationnet} shows a schematic drawing of the architecture of the employed SegmentationNet. It consists of two parts, i.e., the encoder part (contracting path) and the decoder part (expansive path). The encoder part focuses on analysis and feature representation learning from the input data while the decoder part generates segmentation results, relying on the learned features from the encoder part. Shortcut connections are established between layers of equal resolution in the encoder and decoder paths. 

Previous studies show small convolutional kernels are more beneficial for training and performance \cite{vgg16}. For our SegmentationNet, all convolutional layers use kernel size of $3 \times 3 \times 3$ and strides of 1 and all max pooling layers uses kernel size of $2 \times 2 \times 2$ and strides of 2. In the convolutional and deconvolutional blocks of our network,  Batch normalization and Rectified linear unit are adopted.

\textbf{Data augmentation and training patch generation.} For each training sample, besides gray value augmentation and applying a smooth dense deformation field on both data and ground truth labels as suggested in \cite{3DUNET_2016}, we also conduct ROI augmentation following the suggestion in \cite{Sekuboyina_arXiv_2017}. After data augmentation, multiple 3D patches with a size of $160 \times 128 \times 96$ at random locations are taken from each image and saved in a list. The patches in the list are shuffled and packed into batches to fed into the SegmentationNet. 

\textbf{Training.} The training of the SegmentationNet is done in two steps. First, the SegmentationNet is trained for binary segmentation of the lumbar spine, where all vertebra have the same label. The SegmentationNet has to distinguish between spine and background. After that, it is trained for multi-class segmentation where one class corresponds to one vertebra, starting with L1. The trained weights of the binary segmentation network are used to initialize the multi-class segmentation. In both steps, weighted cross-entropy loss functions were optimized with the Adam solver, where considering the fact that we have more background class voxels than voxels of other classes, we reduce the weights of the background voxels and increases the weights of the vertebra voxels to balance the influence \cite{3DUNET_2016}.

\textbf{Testing.} Our trained models can estimate labels of an arbitrary-sized volumetric image. Given a test volumetric image, we extracted overlapped sub-volume patches with the size of $160 \times 128 \times 96$, and fed them to the trained network to get prediction probability maps. For the overlapped voxels, the final probability maps would be the average of the probability maps of the overlapped patches, which were then used to derive the final segmentation results. After that, we conducted morphological operations to remove isolated small volumes and internal holes.

\subsubsection{Implementation Details}
The proposed work was implemented in python using Keras taking Tensorflow as the backend and trained on a desktop with a 3.6GHz Intel(R) i7 CPU and a GTX 1070 graphics card with 8GB GPU memory.

%===========================================================================
\section{Experimental Results}
\label{sec:Exp}
%===========================================================================
\subsection{Experimental design}
We obtained 15 spine CT images with ground truth segmentation from the MICCAI 2016 xVertSeg challenge \footnote{\label{1}One can find details about the xVertSeg challenge at: http://lit.fe.uni-lj.si/xVertSeg/}. Not only are the data acquired with varying FOVs but also contain fractured vertebrae, which posts a challenge for automatic localization and segmentation. In this paper, we conducted a leave-three-out cross-validation study to evaluate the performance of the present method. More specifically, each time we randomly take 3 out of the 15 CT data as the test data and the remaining 12 CT data as the training data. The process was repeated for 5 folds. In each fold, the segmented results of the test data were compared with the associated ground truth segmentation. For each vertebra in a test CT data, we evaluate the Average Symmetric Surface Distance (ASSD) and Hausdorff Distance (HD) between the surface models extracted from different segmentation as well as the volume overlap measurements including Dice Coefficient (DC) and Jaccard Coefficient (JC).

\subsection{Experimental results}
Every time, the SegmentationNet can achieve successful segmentation from the ROI estimated form our LocalizationNet. Quantitative segmentation results of the cross validation study is shown in Table \ref{tab:results_seg_x}, where the results on each individual vertebra as well as on the entire lumbar region are presented. Our approach achieves a mean DC of 95.77$\pm$0.81\% and a mean ASSD of 0.37$\pm$0.06mm on the entire lumbar region.

In each fold, it took about 12 hours and 40 minutes to finish the training for the localization net, and another 23 hours to finish the training of segmentation net. After training, it took on average about 79 seconds to finish the segmentation of one test CT data.

\begin{table}[t]
	\centering
	\caption{Segmentation results of the 5-fold cross validation on the xVertSeg challenge dataset.}
	\label{tab:results_seg_x}
	\resizebox{8.6cm}{!}{%
	\begin{tabular}{c|c|c|c|c|}
		\cline{2-5}
		{\bf } & {\bf DC (\%)} & {\bf JC (\%)} & {\bf HD (mm)} & {\bf ASSD (mm)} \\ \hline
		\multicolumn{1}{|c|}{L1} & $ 96.14\pm0.50 $ & $ 92.56\pm0.91 $ & $ 2.97\pm1.63 $ & $ 0.34\pm0.03 $\\ \hline
		\multicolumn{1}{|c|}{L2} & $ 95.83\pm0.72 $ & $ 92.01\pm1.32 $ & $ 4.38\pm2.43 $ & $ 0.35\pm0.06 $\\ \hline
		\multicolumn{1}{|c|}{L3} & $ 95.89\pm0.63 $ & $ 92.11\pm1.15 $ & $ 4.64\pm3.03 $ & $ 0.36\pm0.04 $\\ \hline
		\multicolumn{1}{|c|}{L4} & $ 95.35\pm0.89 $ & $ 91.12\pm1.61 $ & $ 5.30\pm3.33 $ & $ 0.41\pm0.05 $\\ \hline
		\multicolumn{1}{|c|}{L5} & $ 95.66\pm1.01 $ & $ 91.69\pm1.80 $ & $ 4.34\pm1.44 $ & $ 0.40\pm0.07 $\\ \hline
		\multicolumn{1}{|c|}{Lumbar} & $ 95.77\pm0.81 $ & $ 91.90\pm1.48 $ & $ 4.32\pm2.60 $ & $ 0.37\pm0.06 $\\ \hline
	\end{tabular}
	}
\end{table}

%===========================================================================
\section{Discussions and Conclusions}
\label{sec:Summary}
%===========================================================================
In this paper, we proposed a method based on cascaded FCNs consisting of a localization net and a segmentation net for automatic segmentation of lumbar vertebras from spinal CT data. The localization net helps to turn the attention of our segmentation net to the lumbar region in order to get accurate segmentation of each individual lumbar vertebra. 

The results achieved by our method are comparable to those by the state-of-the-art methods, though direct comparison of different methods is difficult as not all of them are evaluated on the same dataset. For example, when evaluating their method on 50 healthy lumbar vertebrae from 10 spinal CT data, Ibragimov et al. \cite{Ibragimov2014} reported a mean DC of 93.6\%. On the same lumbar vertebral dataset, Korez et al. \cite{comp_korez2} reported a mean DC of 95.3\%. In contrast, even evaluated on a challenging dataset with fractured vertebrae and varying FOVs, our method achieved a mean DC of 95.77\%. In comparison to the method introduced in \cite{Sekuboyina_arXiv_2017}, our approach also achieved superior results. A mean DC of 92.7\% was reported in \cite{Sekuboyina_arXiv_2017} while our approach achieved a mean DC of 95.77\%. As both methods are based on deep learning techniques, one possible explanation is that they have used different localization (MLP vs. our 3D regression FCN) and segmentation (2D U-net like FCN vs. our 3D segmentation FCN) methods from ours.

In conclusion, we presented a cascaded CNN-based approach for fully automatic segmentation of lumbar vertebrae from 3D CT data. Our method achieved equivalent or superior results over the state-of-the-art methods.

%\bibliographystyle{IEEEbib}


\begin{thebibliography}{10}

\bibitem{sports_medicine}
{Balmain Sports Medicine},
\newblock ``Common lumbar spine injuries,''
  https://www.balmainsportsmed.com.au/injury-library/injury-information/lumber-and-spine.html,
  2014,
\newblock [Online; accessed 09-10-2017].

\bibitem{Emch_SR_2011}
T.~Emch and M.~Modic,
\newblock ``Imaging of lumbar degenerative disk disease: history and current
  state.,''
\newblock {\em Skeletal Radiology}, vol. 40, no. 9, pp. 1175--1189, 2011.

\bibitem{Parizel_EUJ_2010}
P.M. Parizel, T.~van~der Zijden, S.~Gaudino, M.~Spaepen, M.H.J. Voormolen, and
  et~al.,
\newblock ``Trauma of the spine and spinal cord: imaging strategies,''
\newblock {\em Eur Spine J}, vol. 19, no. Suppl 1, pp. S8--S17, 2010.

\bibitem{Chu_PLOSONE_2015}
C.~Chu, D.L. Belavy, G.~Armbrecht, M.~Bansmann, D.~Felsenberg, and G.~Zheng,
\newblock ``Fully automatic localization and segmentation of 3d vertebral
  bodies from ct/mr images via a learning-based method,''
\newblock {\em PLOS ONE}, vol. 10, no. 11, pp. e0143327, 2015.

\bibitem{Yao_CMIG_2016}
J~Yao, J.E. Burns, D.~Forsberg, and et~al.,
\newblock ``A multi-center milestone study of clinical vertebral ct
  segmentation,''
\newblock {\em Computerized Medical Imaging and Graphics}, vol. 49, pp. 16--28,
  2016.

\bibitem{Klinder2009}
T.~Klinder, J.~Ostermann, M.~Ehm an~A.~Franz, R.~Kneser, and C.~Lorenz,
\newblock ``Automated model-based vertebra detection, identification, and
  segmentation in ct image,''
\newblock {\em Medical Image Analysis}, vol. 13, no. 3, pp. 471--482, 2009.

\bibitem{Stern2011}
D.~$\check{S}$tern, B.~Likar, F.~Pernus, and T.~Vrtovec,
\newblock ``Parametric modelling and segmentation of vertebral bodies in 3d ct
  and mr spine images,''
\newblock {\em Phys. Med. Biol}, vol. 56, pp. 7505--7522, 2011.

\bibitem{Ibragimov2014}
B.~Ibragimov, B.~Likar, F.~PERNUS, and T.~Vrtovec,
\newblock ``Shape representation for efficient landmark-based segmentation in
  3-d,''
\newblock {\em Medical Imaging, IEEE Transactions on}, vol. 33, no. 4, pp.
  861--874, April 2014.

\bibitem{comp_korez2}
R~Korez, B~Ibragimov, B~Likar, F~Pernus, and T~Vrtovec,
\newblock ``A framework for automated spine and vertebrae interpolation-based
  detection and model-based segmentation,''
\newblock {\em IEEE Trans Med Imaging}, vol. 34, pp. 1649--1662, 01 2015.

\bibitem{MichaelKelm_MedIA_2013}
B.~Michael~Kelm, M.~Wels, S.~Zhou, S.~Seifert, M.~Suehling, Y.~Zheng, and
  D.~Comaniciu,
\newblock ``Spine detection in ct and mr using iterative marginal space
  learning,''
\newblock {\em Medical Image Analysis}, vol. 17, no. 8, pp. 1283--1292, 2013.

\bibitem{Zhan_MICCAI_2012}
Y.~Zhan, D.~Maneesh, M.~Harder, and X.~Zhou,
\newblock ``Robust mr spine detection us- ing hierarchical learning and local
  articulated model.,''
\newblock in {\em Proceedings of MICCAI 2012}, 2012, pp. 141--148.

\bibitem{Glocker_MICCAI_2012}
B.~Glocker, J.~Feulner, and et~al.,
\newblock ``Automatic localization and identication of vertebrae in arbitrary
  field-of-view ct scans�,''
\newblock in {\em Proceedings of MICCAI 2012}, 2012, pp. 590--598.

\bibitem{long2015fully}
Jonathan Long, Evan Shelhamer, and Trevor Darrell,
\newblock ``Fully convolutional networks for semantic segmentation,''
\newblock in {\em Proceedings of the IEEE Conference on Computer Vision and
  Pattern Recognition}, 2015, pp. 3431--3440.

\bibitem{ronneberger2015u}
Olaf Ronneberger, Philipp Fischer, and Thomas Brox,
\newblock ``U-net: Convolutional networks for biomedical image segmentation,''
\newblock in {\em International Conference on Medical Image Computing and
  Computer-Assisted Intervention}. Springer, 2015, pp. 234--241.

\bibitem{3DUNET_2016}
O.~Cicek, A.~Abdulkadir, S.~Lienkamp, T.~Brox, and O.~Ronneberger,
\newblock ``3d u-net: Learning dense volumetric segmentation from sparse
  annotation,''
\newblock in {\em MICCAI 2016}, vol. LNCS 9901, pp. 424--432. Springer, 2016.

\bibitem{Chen_MICCAI_2015}
H.~Chen, C.~Shen, J.~Qin, and et~al.,
\newblock ``Automatic localization and identification of vertebrae in spine ct
  via a joint learning model with deep neural network,''
\newblock in {\em Proceedings of MICCAI 2015}, 2015, pp. 515--522.

\bibitem{Suzani_MICCAI_2015}
A.~Suzani, A.~Seitel, Y.~Liu, and et~al.,
\newblock ``Fast automatic vertebrae detection and localization in pathological
  ct scans - a deep learning approach,''
\newblock in {\em Proceedings of MICCAI 2015}, 2015, pp. 678--786.

\bibitem{Sekuboyina_arXiv_2017}
A.~Sekuboyina, A.~Valentinitsch, J.S. Kirschke, and Menze B.H.,
\newblock ``A localisation-segmentation approach for multi-label annotation of
  lumbar vertebrae using deep nets,''
\newblock {\em arXiv:1703.04347}, 2017.

\bibitem{ioffe2015batch}
S.~Ioffe and C.~Szegedy,
\newblock ``Batch normalization: Accelerating deep network training by reducing
  internal covariate shift,''
\newblock in {\em Proceedings of ICML}, 2015, pp. 448--456.

\bibitem{unitbox}
Jiahui Yu, Yuning Jiang, Zhangyang Wang, Zhimin Cao, and Thomas~S. Huang,
\newblock ``Unitbox: An advanced object detection network,''
\newblock {\em CoRR}, vol. abs/1608.01471, 2016.

\bibitem{KDE_AS_2010}
Z.~Botev, J.~Grotowski, D.~Kroese, and et~al,
\newblock ``Kernel density estimation via diffusion,''
\newblock {\em The Annals of Statistics}, vol. 38, no. 5, pp. 2916–2957,
  2010.

\bibitem{vgg16}
Karen Simonyan and Andrew Zisserman,
\newblock ``Very deep convolutional networks for large-scale image
  recognition,''
\newblock {\em CoRR}, vol. abs/1409.1556, 2014.

\end{thebibliography}
\end{document}